%% file: 00_main.tex
\documentclass[letterpaper, 10 pt, journal, twoside]{IEEEtran}

\ifCLASSINFOpdf
  \usepackage[pdftex]{graphicx}
  \graphicspath{{figures/}}
\else
  \graphicspath{{../eps/}}
\fi

\usepackage{textcomp}
\usepackage{gensymb} 
\usepackage{xcolor}
\usepackage{enumitem} 

\usepackage{hyperref} 

\usepackage{xspace} 
\usepackage{verbatim} 
\usepackage{multirow}
\usepackage{booktabs}    
\usepackage{array}       
\usepackage{adjustbox}   
\usepackage{makecell}  
\newcommand{\systemName}{\textsc{Rampa}\xspace}

\ifCLASSOPTIONcompsoc
 \usepackage[caption=false,font=normalsize,labelfont=sf,textfont=sf]{subfig}
\else
 \usepackage[caption=false,font=footnotesize]{subfig}
\fi

\usepackage{float}
\usepackage{booktabs}

\begin{document}

\title{\systemName: \textbf{R}obotic \textbf{A}ugmented Reality for \textbf{M}achine
\textbf{P}rogramming by Demonstr\textbf{A}tion}

\author{Fatih Dogangun$^{1}$, Serdar Bahar$^{1}$, Yigit Yildirim$^{1}$, Bora Toprak Temir$^{1}$, Emre Ugur$^{1}$, and Mustafa Doga Dogan$^{2,1}$%
\thanks{Manuscript received: October, 12, 2024; Revised January, 14, 2025; Accepted February, 15, 2025.}
\thanks{This paper was recommended for publication by Editor Jee-Hwan Ryu upon evaluation of the Associate Editor and Reviewers' comments. This work has been supported by the \textit{INVERSE} project (no. 101136067), funded by the EU.} 
\thanks{$^{1}$ Department of Computer Engineering, Bogazici University, Istanbul, Türkiye. $^{2}$ Adobe Research, Basel, Switzerland. We thank Muhammet Batuhan Ilhan and Emre Batuhan Goc for their contributions to the initial implementation of the \systemName system. Corresponding author: {\tt\scriptsize fatih.dogangun@std.bogazici.edu.tr}}
\thanks{Code is available at {\tt\scriptsize\url{https://github.com/dogadogan/rampa}}}
\thanks{Digital Object Identifier (DOI): see top of this page.}
}

\markboth{IEEE Robotics and Automation Letters. Preprint Version. Accepted February, 2025}
{Dogangun \MakeLowercase{\textit{et al.}}: \systemName: \textbf{R}obotic \textbf{A}ugmented Reality for \textbf{M}achine
\textbf{P}rogramming by Demonstr\textbf{A}tion}

\maketitle

\begin{abstract}
This paper introduces 
Robotic Augmented Reality for Machine Programming by Demonstration (\systemName), the first ML-integrated, XR-driven end-to-end robotic system, allowing training and deployment of ML models such as ProMPs on the fly, and utilizing the capabilities of state-of-the-art and commercially available AR headsets, e.g., \textit{Meta Quest 3}, to facilitate the application of Programming by Demonstration (PbD) approaches on industrial robotic arms, e.g., \textit{Universal Robots UR10}.
Our approach enables \textbf{\textit{in-situ}} data recording, visualization, and fine-tuning of skill demonstrations directly within the user's physical environment. \systemName addresses critical challenges of PbD, such as safety concerns, programming barriers, and the inefficiency of collecting demonstrations on the actual hardware.
The performance of our system is evaluated against the traditional method of kinesthetic control in teaching three different robotic manipulation tasks and analyzed with quantitative metrics, measuring task performance and completion time, trajectory smoothness, system usability, user experience, and task load using standardized surveys.
Our findings indicate a substantial advancement in how robotic tasks are taught and
refined, promising improvements in operational safety, efficiency, and user engagement in robotic programming.

\end{abstract}


%
\IEEEpeerreviewmaketitle

\input{01_introduction}

\input{02_relatedwork}
\input{03_system}
\input{04_evaluation}

\input{05_limitations}
\input{06_conclusion}


%



\ifCLASSOPTIONcaptionsoff
  \newpage
\fi

\bibliographystyle{IEEEtran}
\bibliography{ARobot}

\end{document}

%% file: 01_introduction.tex
\section{Introduction}

The advent of sophisticated robotic applications necessitates advanced programming techniques that can accommodate complex, dynamic environments. The conventional approach to robot programming typically involves \textit{segmented} processes where users must \textit{alternate between physical and digital realms}.
Traditional approaches rely heavily on offline simulation and manual coding, which can be time-consuming and often impractical for real-time applications \cite{ravichandar2020recent}.
As these methods lack the capability to adapt quickly to new environments or to intuitively incorporate human expertise, the complexity and time involved in programming increases, which also increases the risk of errors and accidents.

The procedure known as Programming by Demonstration (PbD), a.k.a. Learning from Demonstration (LfD), is a widely adopted procedure that enables robots to learn skill policies by watching an expert \cite{argall2009survey}. Recently, Machine Learning (ML)-based PbD frameworks have become instrumental in enabling robots to model and execute intricate skills autonomously. Yet, integrating such approaches into robotic programming often encounters significant challenges, including the labor-intensive and potentially hazardous processes involved in data gathering and model testing in the real world ~\cite{suzuki2022augmented}.

\begin{figure}[t]
\centerline{\includegraphics[width=0.4\textwidth]{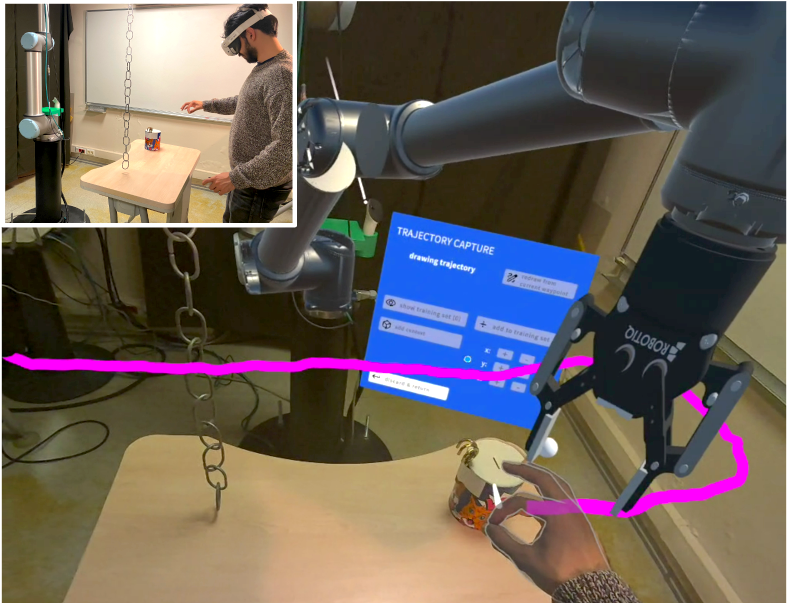}}
\caption{With \systemName, novice and expert users can demonstrate trajectories and simulate robot movements \textit{in situ} using augmented reality (AR).}
\label{fig:fig_insitu}
\end{figure}

In response to these challenges, we introduce a sophisticated AR-based framework that transforms robotic programming through direct, interactive modeling and visualization. Our system, Robotic Augmented Reality for Machine Programming by Demonstration (\systemName), leverages the immersive and interactive capabilities of state-of-the-art XR headsets (e.g., \textit{Meta Quest 3}) to create a cohesive and safe user experience for robot programming and automation of tasks \textbf{\textit{in situ}}. As shown in Fig. \ref{fig:fig_insitu}, users can directly demonstrate robot trajectories and visualize robot movements 
while maintaining the awareness of the actual workspace. We also release \systemName as open source for the robotics community, offering compatibility with cutting-edge mixed reality headsets and robotic arms.


By integrating in-situ simulation and direct manipulation capabilities, \systemName offers a dual advantage: it streamlines the programming process while enhancing the physical safety, flexibility of robotic operations, and efficiency of PbD procedures and, therefore, reduces the barriers to effective human-robot collaboration. The ability to produce and adjust demonstration trajectories in real-time and in situ facilitates the task of the operator while interacting directly with the learned models establishes a deeper understanding of these models.

Our system facilitates a seamless, intuitive interface for:
\paragraph{Data Recording}
Operators use XR to demonstrate skills, capturing the data directly in the user's environment. Compared to traditional methods, such as kinesthetic control, \systemName simplifies the collection of demonstration trajectories, an integral component of PbD procedures, offering a more efficient experience for robotic programmers.

\begin{figure}[t!]
    \begin{center}
        \includegraphics[width=\columnwidth]{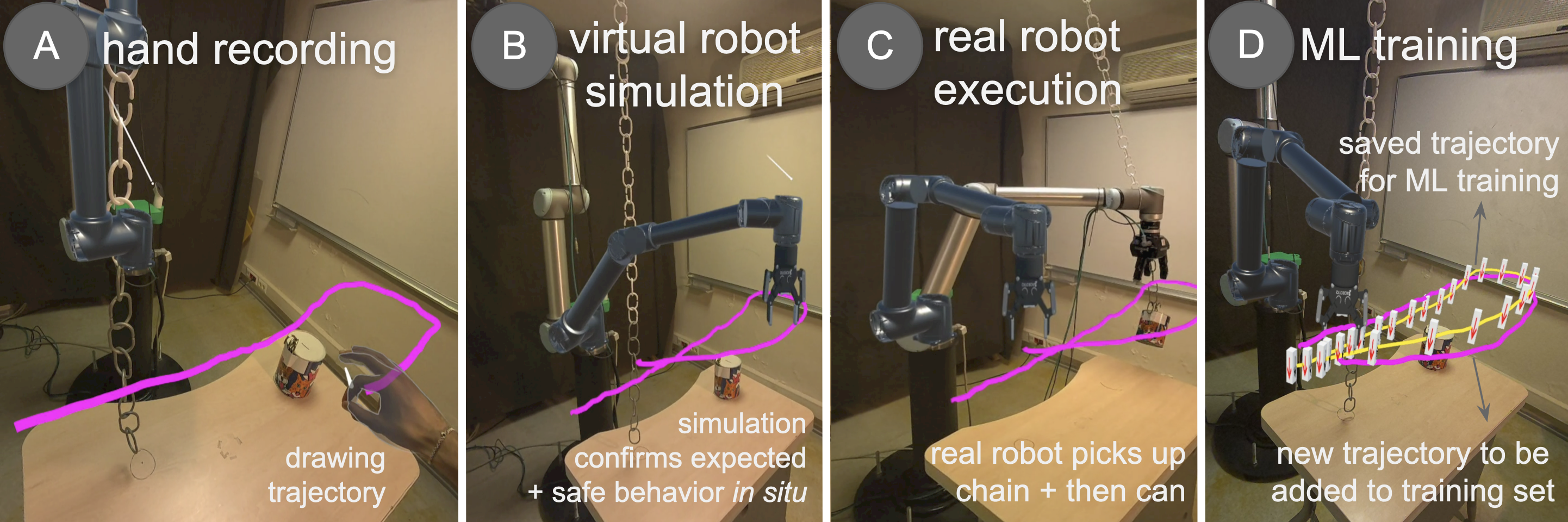}
        \vspace{-0.3cm}
        \caption{\systemName allows users to (a) draw trajectories with their bare hand, (b) simulate and observe their effect \textit{in situ}, and (c) execute them on the real robot. (d) \systemName facilitates the direct and streamlined collection of trajectory data to train ML models. Please refer to text for use cases of the system.} 
        \label{fig:main_figure}
        \vspace{-0.2cm}
    \end{center}
    
\end{figure}

\paragraph{Data Playback and Adjustment}
The system allows operators to visualize and modify the movements of a virtual robot before these movements are executed on the actual robotic arm. This step ensures that potential errors can be corrected in a safe, virtual space. Moreover, the ability to adjust any portion of demonstration trajectories increases the operator's overall performance.

\paragraph{Model Training and Validation}
Utilizing the collected data, popular PbD models, such as Probabilistic Movement Primitives (ProMPs) \cite{paraschos2013probabilistic}, can be trained and validated within the AR environment. Operators can observe the generated trajectories in real-time, ensuring the robot's behavior aligns with desired outcomes.

To the best of our knowledge, we propose the first end-to-end XR-driven PbD framework to enable the demonstration of robotic trajectories, training, and testing ML model augmented by visualization of the robot motion and trajectory adjustment in situ. An overview of the proposed system can be seen in Fig. \ref{fig:main_figure}. Importantly, our comprehensive evaluation comprises the experiment consisting of three distinct robotic tasks, 20 participants including both experienced and novice users, and an extensive analysis using quantitative metrics, incorporating standard surveys, such as \textit{System Usability Scale} (SUS)~\cite{brooke1996sus}, \textit{User Experience Questionnaire}  (UEQ)~\cite{schrepp2017design}, and \textit{NASA Task Load Index} (NASA-TLX)~\cite{hart1988development}, and qualitative user feedback.

This paper details the development and implementation of our system, evaluates its performance against the traditional method of kinesthetic control, and demonstrates its effectiveness through empirical studies. Our findings indicate a significant advancement in how robotic tasks are taught and refined, promising notable improvements in operational safety, efficiency, and user engagement in robotic programming.

In summary, we make the following contributions:
\begin{itemize}
    \item A novel AR-based system that enables users to create and fine-tune robotic applications that mimic human behavior, allowing safe, in-situ simulation and execution.
    \item Comprehensive evaluation and analysis of the system in teaching different robot manipulation tasks using both qualitative feedback and quantitative metrics measuring task performance and completion time, trajectory smoothness, system usability, user experience, and task load.
    \item Facilitation of interactive and situated programming of PbD models suitable for commonly used movement primitives in robotics.
\end{itemize}

%% file: 02_relatedwork.tex
\section{Related Work}
\subsection{Movement Primitives in Programming from Demonstration}

Movement primitives (MP), which provide effective abstractions that facilitate motor control \cite{schaal2005learning}, is a modular formalism widely employed to encode and reproduce complex movements. Schaal et al.~\cite{schaal2005learning} proposed the Dynamical Movement Primitives (DMPs) model to learn MPs by leveraging human expertise. The original DMP model could encode only a single trajectory, while different demonstrations may contain important variabilities that the modeling framework must capture \cite{saveriano2023dynamic,ugur2020compliant}. In order to capture the variety in demonstrations, Gribovskaya et al.~\cite{ gribovskaya2011learning} proposed to model movement trajectories using probabilistic methods. These methods allowed the use of multiple demonstrations to model the skill trajectories, enhancing the representational power and the generalization capacity of MPs~\cite{yin2014learning}.

Latest advances in ML have significantly influenced PbD approaches, enabling the synthesis of more flexible and generalizable MP-based controllers to perform tasks with higher efficiency and adaptability. Recent studies, such as P\'erez-Dattari et al.~\cite{perez2023stable}, Blessing et al.~\cite{ blessing2024information}, Yildirim et al.~\cite{ yildirim2024conditional}, use deep neural networks to model movement trajectories in the form of MPs. However, they are not practical for real-time applications, specifically for human-in-the-loop interactive systems, as they need a training phase to model demonstration trajectories. Due to the real-time requirement, ProMPs, which models a distribution of trajectories as a probability distribution over a set of basis functions, improving the generalization capability to novel via points \cite{paraschos2018using}, is used as the MP modeling framework in this study.

In \systemName, integrating AR into the ML training and visualization process allows users to demonstrate tasks within an AR environment, eliminating manual spatial data recording. Moreover, \systemName enhances the comprehensibility of ML models by offering a interactive platform for ML training and fine-tuning.

\subsection{AR-Oriented Interfaces}
 
Different types of robot programming frameworks, such as visual programming methods Paxton et al. ~\cite{paxton2017costar} combined with speech-based programming Alexandrova et al.~\cite{alexandrova2014robot} have been proposed to allow users to communicate instructions and feedback to robots with ease and high precision, which improves the functionality and usability of robotic systems and enables efficient human-robot interaction. Recently, foundation models have been deployed in real-world robotic applications; nevertheless, they are still inadequate for tasks requiring human involvement~\cite{kawaharazuka2024real}.

Unlike the traditional methods, AR technologies, with their real-time and contextual feedback directly within the user's field of view, make it possible to have more natural and efficient human-robot interactions. As a subset of XR technologies, they enable the superimposition of computer graphics onto the physical world~\cite{green2008human} and have been increasingly utilized to enhance human-robot interaction (HRI), bridging the gap between virtual and real environments. 
Recent surveys Walker et al.~\cite{walker2023virtual} and Suzuki et al.~\cite{suzuki2022augmented} demonstrate the potential of AR to transform robot programming, providing a detailed taxonomy of AR applications in robotics, including beneficial aspects of AR for robot control, programming, and visualization. Specifically, Jiang et al.~\cite{Jiang} compare and contrast multiple interaction interfaces for AR-based robot programming, highlighting how different interface designs impact data collection efficiency and overall user experience.

AR visualization of the robot's motion has been used to simplify the process of planning, observing, and modifying the robot's trajectory for users. For instance, Cao et al.~\cite{cao2019ghostar} enables users to visualize and adjust robot trajectories in real-time, which improves the accuracy of robot actions in human-robot collaborative operations. Similarly, Quintero et al. ~\cite{quintero2018robot} introduces a method that allows users to interact with robot trajectories with ease by using an AR interface to visualize the motion of the robot. Walker et al.~\cite{walker2018communicating}, Rosen et al.~\cite{rosen2020communicating} show that visualizing robot movement improves task efficiency and enhances user understanding of robot intentions, hence facilitating human-robot interactions.

AR interfaces have also been used for robot training methods such as PbD and teleoperation, as they offer interactive and intuitive ways to teach robots new skills.
Arunachalam et al.~\cite{arunachalam2023holo} provides users with a broad AR-based framework for robot learning with teleoperation. Alongside, Luebbers et al. ~\cite{luebbers2021arc} offers in-situ visualizations of robot trajectories with their constraints, and Diehl et al.~\cite{Diehl} proposes a system for visualization of learned skills for verification and error detection, therefore providing an intuitive platform for users to convey proper demonstrations to the robot. However, these studies require kinesthetically demonstrated trajectories. Lotsaris et al.~\cite{Lotsaris} introduces an AR system to demonstrate robot trajectories by placing the virtual gripper in the target position. This work, however, does not allow the user to demonstrate the precise path. Recent works Chen et al.~\cite{chen2024mr} and Chen et al.~\cite{ChenARCap} show the potential of AR interfaces, focusing on the demonstration collection phase. However, they do not offer an end-to-end solution. By bridging the gap left by AR-based interfaces for PbD, we propose a comprehensive, \textit{end-to-end system} that allows users to demonstrate robotic trajectory, visualize the robot motion, modify the trajectory, train, and test the ML model \textit{in situ}, hence making robotic programming more accessible and practical.

A comprehensive comparison with existing AR-based systems is provided in Table \ref{tab:method_comparison}, evaluating several key features. The \textit{End-to-End} feature corresponds to whether the proposed system supports demonstration collection, the training and deployment of the training result on the fly, within a single system. \textit{ML in AR} indicates whether any ML component is integrated into the AR system. \textit{Full Trajectory Modification} refers to whether the system has rewinding, redrawing capabilities of the collected trajectory pre- or post-demonstration by the actual robot. \textit{Online Hand Mimicking by Robot} corresponds to whether the system enables online hand mimicking during demonstration collection by either the simulated or the actual robot. \textit{Trajectory Visualization} represents the capability to visualize the collected trajectory within AR, where \textit{Partial} corresponds to constraint visualization of the learned skill in Luebbers et al. ~\cite{luebbers2021arc}, and end-effector visualization in Lotsaris et al. ~\cite{Lotsaris}. Lastly,  \textit{Trajectory Drawing} describes the method of trajectory collection within AR.

\begin{table}[t]
\centering
\caption{Comparison of AR-Based Systems}
\label{tab:method_comparison}
\renewcommand{\arraystretch}{2.25} 
\resizebox{\columnwidth}{!}{%
\fontsize{50}{50}\selectfont
\begin{tabular}{@{}lcccccc@{}}
\noalign{\hrule height 5pt}
\textbf{Feature / Method} & \textbf{\cite{quintero2018robot}} & \textbf{\cite{luebbers2021arc}} & \textbf{\cite{Lotsaris}} & \textbf{\cite{chen2024mr}} & \textbf{\cite{ChenARCap}} & \textbf{\systemName} \\ \midrule

\bfseries\makecell[l]{End-to-End} & 
No & 
No & 
No & 
No & 
No & 
\textbf{Yes} \\ 

\bfseries\makecell[l]{ML (Training + Deployment) in AR} & 
No & 
\textbf{Yes} & 
No & 
No & 
No & 
\textbf{Yes} \\ 

\bfseries\makecell[l]{Full Trajectory Modification} & 
No & 
No & 
No & 
No & 
No & 
\textbf{Yes} \\ 

\bfseries\makecell[l]{Online Hand Mimicking by Robot} & 
No & 
No & 
No & 
\textbf{Yes} & 
\textbf{Yes} & 
\textbf{Yes} \\ 

\bfseries\makecell[l]{Trajectory Visualization} & 
\textbf{Yes} & 
Partial & 
Partial & 
No & 
\textbf{Yes} & 
\textbf{Yes} \\ 

\bfseries\makecell[l]{Trajectory Drawing Method} & 
Point-based & 
None & 
Point-based & 
\makecell{Hand \\ gestures} & 
\makecell{Hand \\ gestures} & 
\makecell{Hand \\ gestures} \\ 

\noalign{\hrule height 5pt}
\end{tabular}%
}
\end{table}

%% file: 03_system.tex
\section{\systemName: An End-to-End System for Robotic AR}
The \systemName system integrates a ROS server, Unity game engine, and Quest 3 XR headset to provide an intuitive platform for interactive demonstration collection, ML training, and visualization. Its interface, implemented in Unity, include a simulated robot, a UI menu, and interactive tools such as virtual markers for model conditioning. Users control the system via hand gestures and controller inputs, leveraging
features like trajectory adjustment, auto-calibration, online
hand following, and ML integration.

\subsection{System Overview}
\systemName features key functionalities to facilitate safe and cohesive trajectory capturing with hand mimicry, fine-tuning of the demonstrated trajectories, use of this recorded data to train ML models and safe execution of the recorded trajectories in the virtual environment, or the trajectories sampled from the trained model on the real robot.

\subsubsection{Simulated Robot and Auto-Calibration}
A simulated robot is present to model the actual robot’s behavior, allowing for accurate testing and validation of trajectory plans. For deployment of demonstrations to actual robot, the simulated and the actual robot should be precisely aligned. This problem is addressed by the auto-calibration feature. The feature uses the controller from the headset and a 3D-printed tool that can hold the controller for the virtual robot to be aligned with the actual robot, as shown in Fig. \ref{fig:main_figure}-A. The simulation environment ensures that any trajectory can be tested thoroughly for safety before being deployed on the real robot, also making sure the actual robot will replicate both the location and the motion of the simulated one.

\subsubsection{Trajectory Playback and Adjustment}

\begin{figure}[t!]
\centerline{\includegraphics[width=0.49\textwidth]{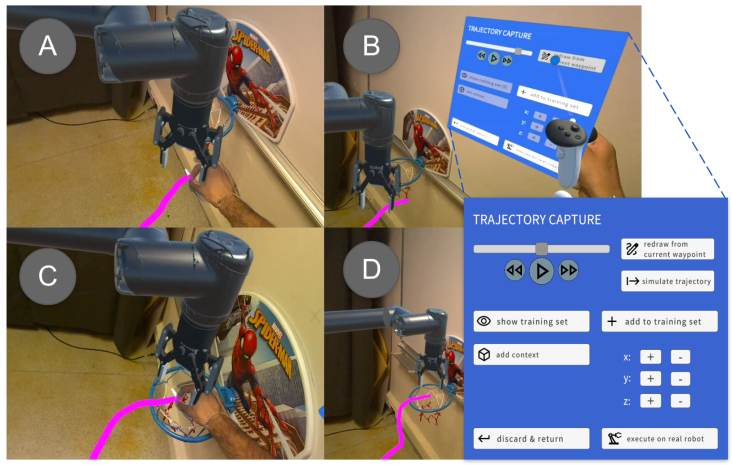}}
\caption{The user (a) draws a trajectory to the hoop, (b) rewinds the trajectory and chooses to redraw it from the current way-point using the UI menu, (c) redraws the trajectory, and (d) finalizes the trajectory.}
\label{fig:fig_traj_adjust}
\end{figure}

Trajectory playback and adjustment are implemented to allow the user to fine-tune the recorded trajectory. When a user decides to inspect the current trajectory with the simulated robot, before or after real robot demonstration, they can go backward or forward in the trajectory, play the rest of the re-winded trajectory, and pause when desired (Fig. \ref{fig:fig_traj_adjust}), using the UI provided in the system. If the user decides to discard a portion of the trajectory, they can redraw the trajectory from the current way-point. This interactive approach enables precise control of the robot's movements and ensures that the final trajectory is both accurate and safe for the desired task.

\subsubsection{Real-Time Hand Following}
The user may prefer having the simulated robot follow the position and orientation of their hand, as shown in Fig. \ref{fig:fig_insitu}. This enables the user to plan more effectively and refine trajectories in a safe and controlled environment, avoiding potential safety issues or collisions. The immediate feedback allows previewing how the robot will navigate through the real objects.

\begin{figure}[b]
\centerline{\includegraphics[width=0.4\textwidth]{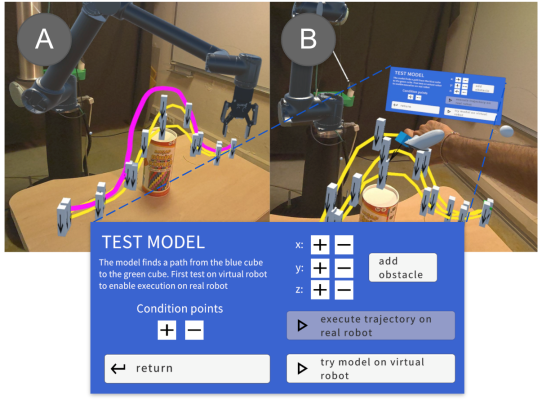}}
\caption{Users can save and view trajectories they have recorded, then train an ML model, condition it using virtual cubes, and generate a new trajectory.}
\label{fig:fig_ml_training}
\end{figure}

\subsubsection{ML Training and Preview}
The recorded trajectories can be either discarded or added to a training set, which can then be viewed or deleted, as shown in Fig. \ref{fig:fig_ml_training}. \systemName allows the user to add demonstrated trajectory to the training set after the simulation of the robot motion and/or physical robot execution. Moreover, the user can modify the recorded trajectory post-simulation and/or post-execution, which enhances the adaptability of the system and flexibility of the user during the demonstration collection process. The training set can then be used to train a model. To test the model, the user can add way-points to condition the trained model using virtual cubes, as demonstrated in Fig. \ref{fig:fig_ml_training}. This process facilitates the tasks of training, conditioning, and sampling trajectories efficiently.

\subsubsection{Execution on the Real Robot}
The execution of the recorded trajectories and the sampled trajectories from the trained model can be performed using the virtual user interface. This capability actualizes the transfer of trained tasks from simulation to real-world execution.

\subsection{Implementation}

\begin{figure}[b!]
\centerline{\includegraphics[width=0.50\textwidth]{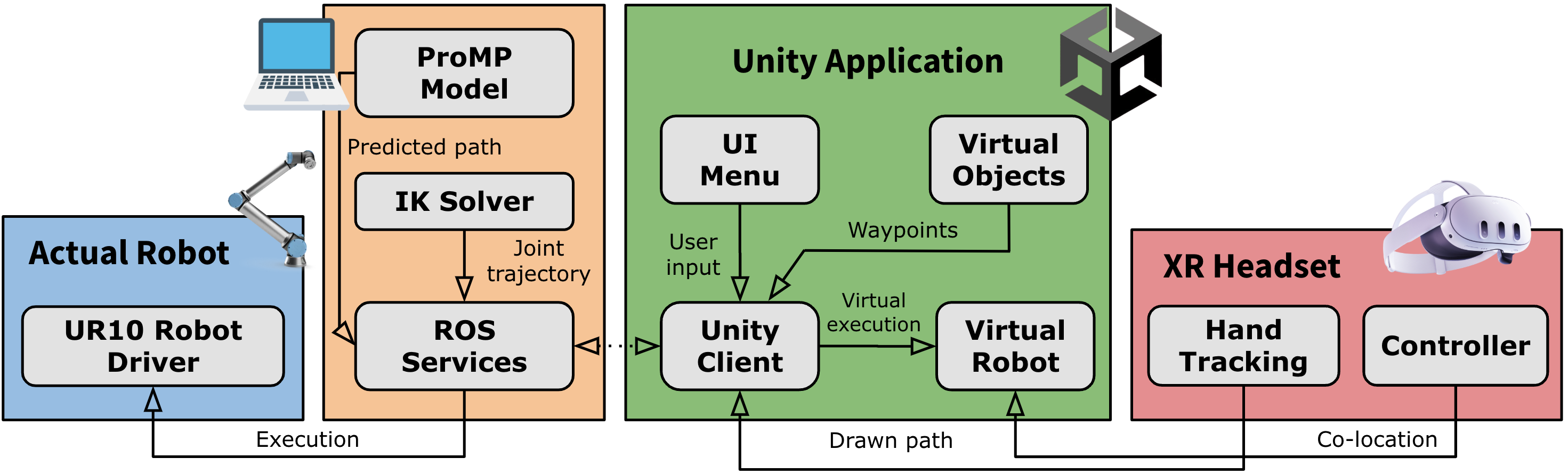}}
\caption{Technical workflow of \systemName.}
\label{fig:fig_workflow}
\end{figure}

The implementation can be divided into four parts: In the game engine, where Unity is used, the main functionalities are implemented using \textit{Meta Quest} API packages. The ROS part manages the interactions between the application and the IK service and the models used for training, along with handling the communication between the application and the actual robot. The actual robot is used to record the demonstrated trajectories and execute the skills effectively upon request. Finally, the XR headset, specifically \textit{Quest 3}, deploys and operates \textit{Unity} application. The following sections examine each part in detail, complemented by an illustration of the technical workflow in Fig. \ref{fig:fig_workflow}.

\subsubsection{Unity}

The simulated robot is integrated into \textit{Unity}, using the \textit{URDF Importer} package of \textit{Unity}, where the simulated robot replicates the kinematics, dynamics, and physical properties according to the Unified Robot Description Format (URDF) description of the robot. A menu is also implemented, enabling the user to draw and adjust trajectories, switch between different trajectory capture modes, and perform the previously mentioned ML tasks. 

Detecting  the hand movements are done using \textit{Meta Quest}'s hand gesture recognition and the related APIs. The user can choose from the UI menu provided before starting to demonstrate the trajectory whether the positions in the trajectory are recorded with a fixed orientation or if the orientation of the user's hand is also captured. The end-effector’s position is determined to be the position of the user hand pinch. If the user chooses to capture orientation, the hand orientation information provided by the related \textit{Meta Quest} API is used, and the palm-facing-down orientation of the hand is mapped to the orientation in which the end-effector’s tool axis aligns with the negative vertical direction. The user can also choose whether the simulated robot replicates the trajectory after the demonstration or the robot follows the hand in real time.

To enable co-location of the simulated robot with the actual robot, the user places the left controller of the headset to a stable 3D-printed tool. After selecting auto-calibration in the UI, the simulated robot teleports to the position of the controller with a relative offset. For our case, we 3D-printed a tool that can stand on the base of \textit{UR10}; however, it is important to note that any fixed, stable object that can hold the controller can be used to auto-calibrate, alongside necessary offset calculations.

If the user prefers real-time hand-following with the simulated robot, an off-the-shelf IK solver is continuously queried to compute the trajectory segment from the previous hand location to the current. This is done as the user continues to draw the trajectory as shown in Fig. \ref{fig:main_figure}. In this study, the TRAC solver~\cite{beeson} is used for IK computations. Obtained trajectory segments are simultaneously sent to \textit{Unity} for illustration purposes via the simulated robot. If the user does not want to activate real-time hand-following, then the entire trajectory data, containing uniformly sampled and time-stamped 6D Cartesian positions, is recorded first and then sent to the ROS client at once. In either case, the trajectory sampling rate is fixed at 200 milliseconds. Immediately after receiving waypoints, the ROS client uses the IK solver to compute the necessary joint angles to execute the trajectory upon the user’s request. During execution, a warning is displayed if the simulated robot collides with any object in the surroundings.

\systemName enables users to replay entire trajectories or readjust portions of any length using the menu illustrated on the right-hand side of Fig. \ref{fig:fig_traj_adjust}. It also allows saving trajectories and viewing waypoints, which are shown with yellow lines and black arrows in Fig. \ref{fig:fig_ml_training}. One of the most crucial features that \systemName provides is that the recorded data can be used directly to train ML-based models. The user can also condition the ML-based trajectory model by directly observing the actual obstacle and placing a virtual marker accordingly, as shown in Fig. \ref{fig:fig_ml_training}. User-specified time-stamps and 6D Cartesian positions of the placed virtual markers are used to condition the model. The output of the conditioned model is then sent to Unity and simulated by the virtual robot for validation purposes.

\subsubsection{ROS Client}

The ROS server remains on standby to accommodate the requests from the \textit{Unity} application. Upon initialization, the application establishes a TCP connection using the \textit{ROS-TCP-Connector} package from \textit{Unity}. Within the server, multiple ROS services are utilized to manage various types of operations initiated by the \textit{Unity} application. These services operate within a single process and require unique message types that the \textit{Unity} client complies with.

Upon the user’s request, the ROS client reformats the recorded data and trains a new model from scratch. Since neural-network-based frameworks require long training procedures, we demonstrate this feature by training ProMPs.

\subsubsection{Actual Robot}

For the execution of the trajectory on the real robot, a TCP connection is set up between the server and the actual robot. By publishing states from the recorded trajectory in the virtual environment, the actual robot follows the published states.

\subsubsection{XR Headset}

The \textit{Unity} application is deployed on the headset for complete operation with AR experience. For the co-location of the simulated robot with the actual robot, the left controller's position and rotation are used with relevant offset values, as illustrated in Fig. \ref{fig:main_figure}-A. 

%% file: 04_evaluation.tex
\section{Evaluation}

We conducted a comprehensive evaluation using both quantitative and qualitative metrics to assess the competence and performance of \systemName and compare it with Kinesthetic Control (KC). These assessments include measuring the task-completion time, evaluating the performance of trained MP, the smoothness of  trajectories, and user feedback concerning usability, user experience, and task load through questionnaires. 

A total of 20 participants (18 male and 2 female, ages 20-36, M=23.5, SD=3.5) were recruited for the user study. Participants include 10 people with experience in robotics, where 7 of them additionally had experience in AR (experts), and 10 people with no prior experience in AR and robotics (novices). Permission for the user study is taken from the Bogazici University (No:2024/22).

We have conducted an experiment including three robotic tasks; both demonstrated using \systemName system (\systemName condition) and the physical robot in Kinesthetic Control (KC) condition. Participants were equipped with the \textit{Quest 3} XR headset. We used the Universal Robot UR10, and the experiment was conducted in a robotics laboratory on campus.

\subsection{Procedure}
We arranged the user study so that half of the participants first used the \systemName system and then moved the robot in KC mode, while other participants performed these tasks in the opposite order. Before participants started performing tasks, we offered an overview and explanation of the procedure and experiment. Specifically, we showed users how they can utilize trajectory playback, edit, save and deletion features of \systemName, if desired or whenever necessary. Regarding KC mode, the users are informed that they will have to demonstrate from scratch, if they desire to modify their demonstrated trajectory. The users are also told that saving and replaying demonstrated trajectory in KC mode will be handled manually by the experimenter. After participants became familiar with using the \systemName system and moving the real robot in KC mode, we asked them to perform \textbf{three main tasks}:

\paragraph{Warm-up Exercise} Users were asked to move and control the physical robot. To familiarize participants with the \systemName system, they were asked to place a 3D printed controller attachment to ensure the real and virtual robots are correctly calibrated position-wise and draw a simple trajectory. 

\paragraph{Basketball Task (Task 1)} Participants were requested to draw a trajectory where when the simulated robot mimics the route drawn, it can pass a ball it holds through a basketball hoop. The robot initially holds a ball and releases it at the end of the route, which is executed externally by the experimenter.

\paragraph{Chain-Hook Task (Task 2)} Inspired by crane operations for logistics in the industry, a chain hanging from the ceiling and an object on the table with an attached hook were prepared. Users were requested to draw a trajectory for the robot to perform a sequence of actions: moving the chain towards the object, connecting it with the hook, lifting the object from the table, and placing the object back on the table.

\paragraph{Obstacle Avoidance Task (Task 3)} Users were requested to train and test a model for an obstacle avoidance task. They were asked to draw trajectories by avoiding the total of 3 objects with changing sizes and train the model using these trajectories. Then, we placed an object of a different size from the previous objects and asked participants to test their model by specifying it via a point above the object. 

We recorded the completion time for each task. In tasks 1 and 2, the duration for demonstrating a single successful robotic trajectory and the number of trials needed was recorded. In task 3, we measured the time required to demonstrate three distinct robot trajectories, and also saved the demonstrated trajectories for subsequent analysis, allowing for the comparison of their smoothness and similarity among trained MPs. For training, ProMP 
is used and the hyperparameter of weights per dimension is set to 20. Upon completion, 
participants were also asked to complete a survey. During the user study, demonstrations discarded due to technical issues (e.g., headset or robot malfunctions) were not considered.

\subsection{Results \& Discussion}
\paragraph{Task Completion Time}

\begin{figure}[b]
\centerline{\includegraphics[width=0.41\textwidth]{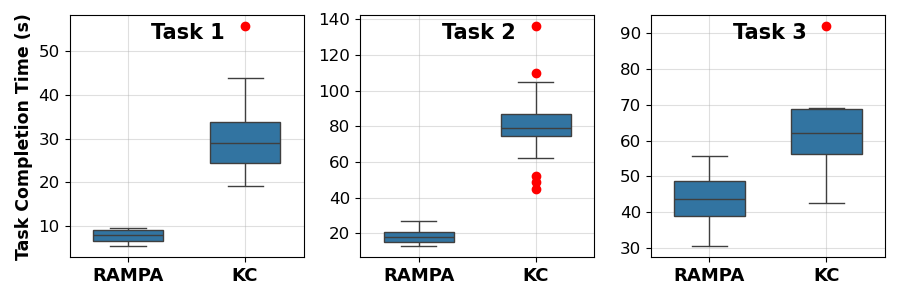}}
\caption{Comparison of task completion times of \systemName and Kinesthetic Control. The box shows the interquartile range (IQR), including the middle 50\% of the data between the first quartile (Q1) and the third quartile (Q3). The whiskers extend from the edges of the box to 1.5 IQR of the Q1 and Q3, and the rest of the data points are considered outliers, shown as red points.}
\label{fig_task_c}
\end{figure}

We measured the task completion time to evaluate the efficiency of the \systemName system compared to KC. Our analysis, as presented in Fig. \ref{fig_task_c}, revealed that the average task completion time is shorter while participants utilize the \systemName system. A paired t-test confirmed significantly shorter completion times in the \systemName condition across all tasks: Task 1 (t = 11.83, p \textless 0.001),  Task 2 (t = 13.93, p \textless 0.001),  Task 3 (t = 7.55, p \textless 0.001).

In the KC condition, Table \ref{tab:task_completion} shows that the task completion time difference between expert and novice users was 22.6\% for Task 1 and 34.5\% for Task 2, indicating a notable difference due to the increased complexity of the demonstration for Task 2 as expected. On the other hand, when utilizing the \systemName system, the task completion time difference was remarkably reduced. Our analysis thus suggests that \systemName allows users to demonstrate robot trajectories efficiently and lessens the impact of prior experience on task performance.

\begin{table}[t!]
\centering
\captionsetup{justification=centering}
\caption{Comparison of Task Completion Times}
\label{tab:task_completion}
\resizebox{0.68\columnwidth}{!}{%
\begin{tabular}{l c c c}\toprule
 &  & \textbf{RAMPA} & \textbf{KC} \\
\midrule
\multirow{2}{*}{\textbf{Task 1}} 
    & Experts &  \multicolumn{1}{c}{\textbf{7.9s $\pm$ 1.3}} & \multicolumn{1}{c}{27.6s $\pm$ 5.2}    \\
    & Novices  & \multicolumn{1}{c}{\textbf{7.5s $\pm$ 1.4}} & \multicolumn{1}{c}{33.3s $\pm$ 10.0} \\
\midrule
\multirow{2}{*}{\textbf{Task 2}} 
    & Experts  & \multicolumn{1}{c}{\textbf{17.7s $\pm$ 3.4}} & \multicolumn{1}{c}{69.5s $\pm$ 15.6} \\
    & Novices  & \multicolumn{1}{c}{\textbf{19.2s $\pm$ 4.8}} & \multicolumn{1}{c}{92.1s $\pm$ 18.9} \\
\midrule
\multirow{2}{*}{\textbf{Task 3}} 
    & Experts  & \multicolumn{1}{c}{\textbf{46.8s $\pm$ 7.0} } & \multicolumn{1}{c}{63.4s $\pm$ 5.3}  \\
    & Novices  & \multicolumn{1}{c}{\textbf{40.9s $\pm$ 7.0} } & \multicolumn{1}{c}{60.7s $\pm$ 13.0} \\
\bottomrule
\end{tabular}%
}
\end{table}

\paragraph{Trained Movement Primitives}

To compare the trajectory generation performances of the models, a ground truth trajectory is required for the new environment where these models are tested. For this, we initially trained a contextual ProMP model using the trajectories demonstrated during Task 3, using the heights of the objects as the model's parameter. We considered the trajectory sampled from this model as the ground-truth trajectory. Next, we computed the Mean Squared Error (MSE) between the ground-truth trajectory and the generated trajectory from the trained and conditioned ProMP by the user during task 3. The result shows that the mean of MSE values is 0.0017 (SD = 0.0024) for the KC and 0.0014 (SD = 0.0011) for \systemName, and indicates that the ProMP model shows comparable performance when trained on trajectories derived using either \systemName or KC. 

\paragraph{Trajectories Smoothness}

We analyzed the smoothness of trajectories recorded using \systemName and KC. The results reveal that the \systemName demonstrations showed a lower jerk (0.004 ${m}/{s^{3}}$ $\pm$ 0.002) compared to the KC demonstrations (0.045 ${m}/{s^{3}}$ $\pm$ 0.011) while exhibiting comparable performance in both the deviation (\systemName: 0.111 ${m}$ $\pm$ 0.018, KC: 0.109 ${m}$ $\pm$ 0.010) and the variation (\systemName: 0.070 ${m^2}$ $\pm$ 0.021, KC: 0.073 ${m^2}$ $\pm$ 0.012), implying superior or comparable smoothness across all metrics. This analysis indicates that \systemName facilitates efficient trajectory demonstration
while maintaining the consistency and quality of demonstrations.

\paragraph{Number of Trials}
All participants successfully completed tasks 1 and 2 in both conditions. For task 1, the number of trials required by each participant was 1, except for one novice user attempting twice while using the \systemName system. For task 2, the average number of trials is 1.111 (SD = 0.314) for expert participants to succeed in both KC and \systemName conditions. For novice users, the mean of the number of trials is 1.2 (SD = 0.4) in KC and 1.4 (SD = 0.49) in \systemName, which is not significantly different (t = -1.5, p = 0.168). A slight increase in the number of trials for task 2 is expected, given the complexity of the demonstration compared to task 1. Our analysis suggests that the number of trials did not show a statistically significant difference between KC and \systemName conditions for both experts and novices in tasks 1 and 2.

\paragraph{Survey Results}

The survey consists of \textit{System Usability Scale} (SUS)~\cite{brooke1996sus}, the shorter version of the \textit{User Experience Questionnaire}  (UEQ)~\cite{schrepp2017design}, the \textit{NASA Task Load Index} (NASA-TLX)~\cite{hart1988development}, and additional questions to compare the different aspects of \systemName and KC. We utilized the SUS to evaluate and compare the usability of \systemName and KC. We obtained a score of 82.75, which shows that the \systemName system is within A grade (i.e., 90\%-95\%) while KC achieved a score of 62.38 which places KC within D grade~\cite{sauro2016quantifying}.

\begin{figure}[t!]
\centerline{\includegraphics[width=0.49\textwidth]{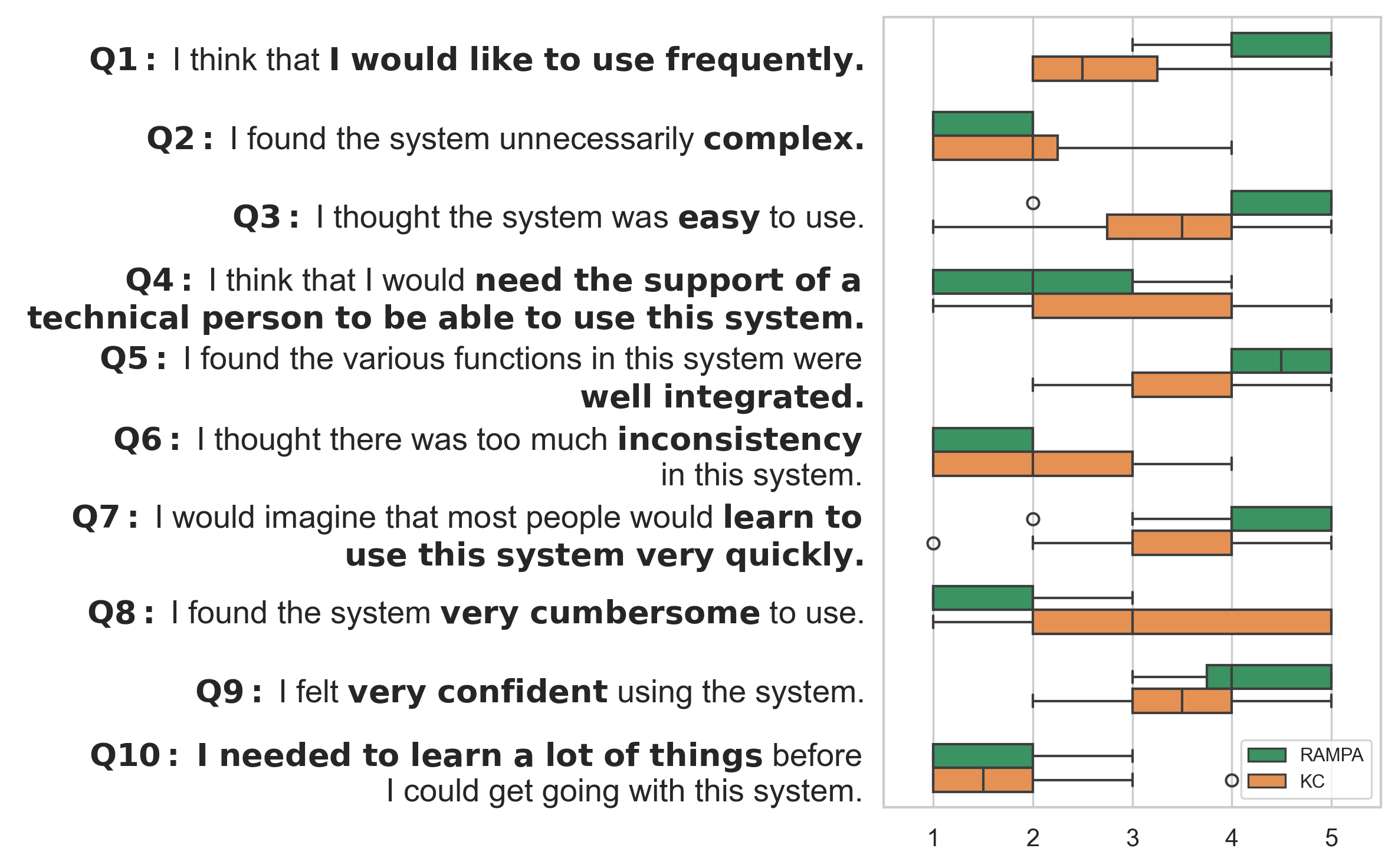}}
\caption{The comparison of SUS results of KC and \systemName conditions.}
\label{fig:fig_sus}
\end{figure}

We employed a subset of questions from the SUS to compare \systemName and KC. As shown in Fig. \ref{fig:fig_sus} , the \systemName system has superior scores across positive statements such as Q1, Q3, Q7, and Q9, which suggests that \systemName provides an easy-to-use framework. The reported scores for Q4, Q7, and Q8 suggest that manually moving the robot might be perceived as more demanding and complex compared to using \systemName. These findings support that \systemName enhances the user experience and satisfaction while simplifying the robotic trajectory demonstration process.

To evaluate the perceived safety of the \systemName system, participants were asked to express which method they found safer. Reported responses were overwhelmingly positive towards \systemName, with 95\% of participants (19 out of 20) indicating that using \systemName is safer than KC. This result shows increased perceived safety with \systemName.

\begin{figure}[b]
\centerline{\includegraphics[width=0.41\textwidth]{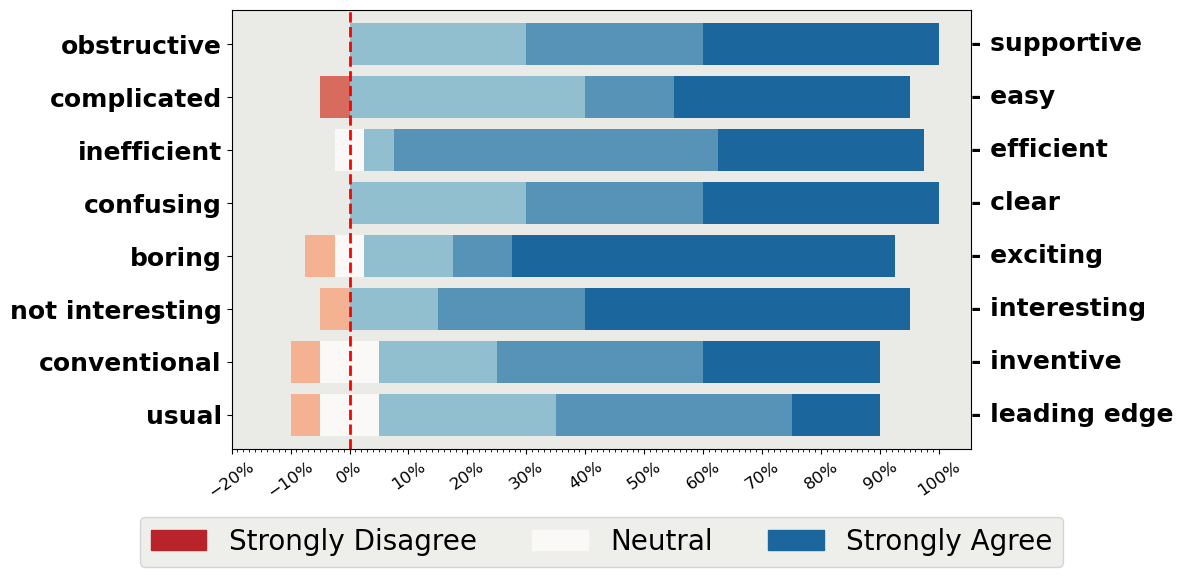}}
\caption{UEQ results of the \systemName condition.}
\label{fig:fig_ueq}
\end{figure}

We employed the shorter version of the UEQ to further evaluate the user's experience of using the \systemName system. As shown in Fig. \ref{fig:fig_ueq}, the \systemName system has positive evaluations across both pragmatic and hedonic qualities. Particularly, \systemName received a score of 2.09 on pragmatic quality, 1.94 on hedonic quality and 2.01 on overall scales, which puts the \systemName system in the range of the top 10\% results \cite{schrepp2017design}.

\begin{figure}[t!]
\centerline{\includegraphics[width=0.41\textwidth]{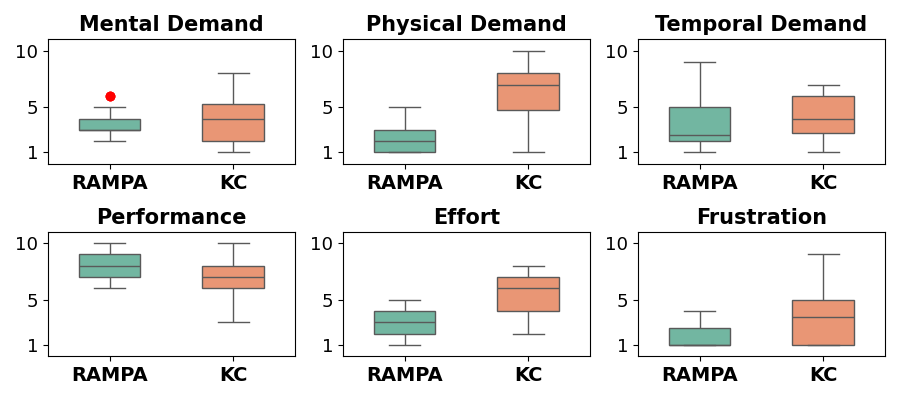}}
\caption{Comparison of NASA-TLX results of the \systemName and KC conditions.}
\label{fig_nasa}
\end{figure}

The analysis of the NASA-TLX survey with a paired t-test reveals that participants using \systemName were more satisfied with their task performance (t = 2.08, p = 0.044) and felt less frustration (t = -2.518, p = 0.016) while indicating that \systemName requires less physical demand (t = -6.92, p \textless 0.001) and effort (t = -5.04, p \textless 0.001) in Fig. \ref{fig_nasa}. In addition, we used the RAW-TLX procedure~\cite{RAW-TLX} to compute aggregated workload scores. The resulting scores are 27.08 for \systemName and 44.33 for KC, indicating that participants generally perceived a lower workload while using the \systemName system compared to KC condition. The increased performance satisfaction with reduced physical demand, effort, and frustration suggests that \systemName provides a user-friendly interface to perform tasks effectively. These findings show that \systemName enhances not only task efficiency but also user experience and comfort.

Participants were asked which approach they would prefer to use for demonstrating robotic trajectory. The reported answers show a significant preference towards \systemName with 95\% (19 out 20 participants), which, considered with previous evaluation results, indicates that \systemName strongly facilitates the efficacy and accessibility of the robot programming process.

\paragraph{Qualitative Feedback}We collected qualitative user feedback from participants to provide a more comprehensive evaluation of \systemName. One participant stated: \textit{“The AR system made me feel safe throughout, and I didn’t have to worry about damaging the robot”} (P20), highlighting the safety advantages of \systemName. Regarding efficiency, one expert participant noted: \textit{“Despite my experience in the domain, it is intimidating even for me to manually control the industrial robots. \systemName simplifies the data collection part of LfD and speeds up my research”} (P11). The system’s intuitiveness was highlighted: \textit{“The controls with hand gestures being straightforward”} (P1) and \textit{“AR integration is done seamlessly. It is easy to understand the trajectory of the robot”} (P19).

\paragraph{Auto-Calibration Accuracy} To assess the auto-calibration method, we conducted an experiment where we moved the robot to a pre-defined point on the table using the \systemName system. Firstly, we established a reference point by moving the robot manually to that point and recording the end-effector position. Then, we performed auto-calibration, demonstrated a trajectory using the \systemName system, and recorded the end-effector position, which was repeated 10 times. The resulting mean of the position error relative to the reference point is 0.0196 m (SD = 0.0134).

%% file: 05_limitations.tex
\section{Limitations}
Our system allows collision detection between the simulated robot and the scanned static environment represented with meshes. It currently does not support visual detection of dynamic objects \cite{dogan2024augmented}, and interaction of the simulated robot with them. For tasks that require intense robot-object interactions, the dynamic objects should be automatically registered as interactable objects to the \textit{Unity Engine} simulator on-the-fly, which requires realistic modeling of all physical properties such as mass, center of mass, detailed shape and density. While functional, auto-calibration might be time-consuming;  easy-to-use solutions like trackable QR codes can facilitate this process~\cite{dogan2022infraredtags}. Moreover, controlling the gripper through the AR interface would improve the completeness of the system, diminishing the need for external operations. Also, a visualization of the robot's workspace can help users grasp its limitations, making the demonstration process less error-prone. \systemName currently advances PbD research by replacing kinesthetic control, lacking a task-centered focus. Future extensions may incorporate egocentric camera views, computer vision, or semantic understanding to mimic daily user interactions and utilize real-world objects. Even though we use the \textit{UR10} robotic platform and ProMPs for convenience, we designed \systemName to be robot- and model-agnostic. The system can be adapted to other robotic hardware if their URDF description and its TRAC IK~\cite{beeson} configuration are provided, necessitating only minimal modifications in the \textit{Unity} application and the ROS client. The system can also be extended by integrating other ML models into the ROS client, if the model can be trained real-time. In addition, although the \textit{Unity} application depends on \textit{Meta Quest} APIs, it can be adapted to any headset, given it has a corresponding XR library with hand recognition.

%% file: 06_conclusion.tex
\section{Conclusion}
We introduced an AR-based framework that provides a comprehensive and intuitive platform for demonstrating robotic skill trajectories, integrating trajectory visualizations and modifications, real-time hand mimicry, and training and testing ML models. These functionalities simplified the demonstration collection in PbD procedures and enhanced the human-robot interaction by leveraging AR. We presented quantitative evaluations, supporting our initial hypothesis of increased effectiveness without sacrificing user experience or task precision. \systemName illustrates the potential to lead to more accessible and efficient robotic programming for both novices and experts.